\documentclass[10pt,twocolumn,letterpaper]{article}

\usepackage{3dv}
\usepackage{times}
\usepackage{epsfig}
\usepackage{graphicx}
\usepackage{amsmath}
\usepackage{amssymb}

\usepackage{color, colortbl}

\usepackage{comment}
\usepackage{color}
\usepackage{booktabs}
\usepackage{multirow}
\usepackage{yfonts}
\usepackage{xspace}
\usepackage{pifont} %
\usepackage{enumitem}

\usepackage{mathrsfs}

\def\sepappendix{0}

\usepackage[pagebackref=true,breaklinks=true,colorlinks,bookmarks=false]{hyperref}

\threedvfinalcopy %

\def\threedvPaperID{23} %

\ifthreedvfinal\fi
\begin{document}

\title{TEACH: Temporal Action Composition for 3D Humans}

\author{Nikos Athanasiou$^{1}$\quad \;
Mathis Petrovich$^{1,2}$\quad \;
Michael J. Black$^{1}$\quad \;
G{\"u}l Varol$^{2}$\\
$^{1}$Max Planck Institute for Intelligent Systems, T{\"u}bingen, Germany\\
$^{2}$LIGM, {\'E}cole des Ponts, Univ Gustave Eiffel, CNRS, France\\
{ \tt\small
    \{nathanasiou,black\}@tue.mpg.de
    \{mathis.petrovich,gul.varol\}@enpc.fr
}
}

\maketitle

\newcommand{\methodname}{TEACH\xspace} %
\newcommand{\figref}[1]{Fig.~\ref{#1}}

\newcommand{\babel}{\texttt{BABEL}\xspace}
\newcommand{\supmat}{Sup.~Mat.\xspace}
\newcommand{\kit}{\texttt{KIT}\xspace}

\renewcommand{\ie}{i.e.\xspace}

\begin{abstract}
Given a series of natural language descriptions, our task is to generate 3D human motions that correspond semantically to the text, and follow the temporal order of the instructions.
In particular, our goal is to enable the synthesis of a series of actions, which we refer to as temporal action composition.
The current state of the art in text-conditioned motion synthesis only takes a single action or a single sentence as input. This is partially due to lack of suitable training data containing action sequences, but also due to the computational complexity of their non-autoregressive model formulation, which does not scale well to long sequences. %
In this work, we address both issues.
First, we exploit the recent BABEL motion-text collection, which has a wide range of labeled actions, many of which occur in a sequence with transitions between them.
Next, we design a Transformer-based approach that operates non-autoregressively within an action, but autoregressively within the sequence of actions. This hierarchical formulation proves effective in our experiments when compared with multiple baselines. %
Our approach, called TEACH for ``TEmporal Action Compositions for Human motions'', produces realistic human motions for a wide variety of actions and temporal compositions from language descriptions.
To encourage work on this new task,
we make our code %
available for research purposes at \url{teach.is.tue.mpg.de}.
\end{abstract}
\section{Introduction}

The  generation of realistic 3D human
motions has applications in virtual reality, the games industry, and any application
that requires motion capture data.
Recently, controlling 3D human motion synthesis with semantics has  received
increasing attention \cite{chuan2020action2motion,ACTOR:ICCV:2021,Ghosh_2021_ICCV,petrovich2022temos}.
The task concerns inputting semantics in the form of categorical actions,
or free-form natural language descriptions, and outputting a series of 3D
body poses. In this work, we address the latter, i.e., text-conditioned motion generation,
which is more flexible compared to pre-defining a set of categories.
More specifically, our goal is to animate a sequence of actions given a sequence of textual prompts.
Generating such \textit{temporal action compositions} has not previously been studied.

Humans move in complex ways that involve different simultaneous and/or sequential actions. Hence, {\em compositionality} in time must be modeled 
to generate everyday motions that contain a series of different actions and that last longer than a few seconds.
Compositionality in space, i.e., simultaneous actions, is another interesting
direction outside the scope of this work.
\begin{figure}[t]
	\centering
	\includegraphics[width=\linewidth]{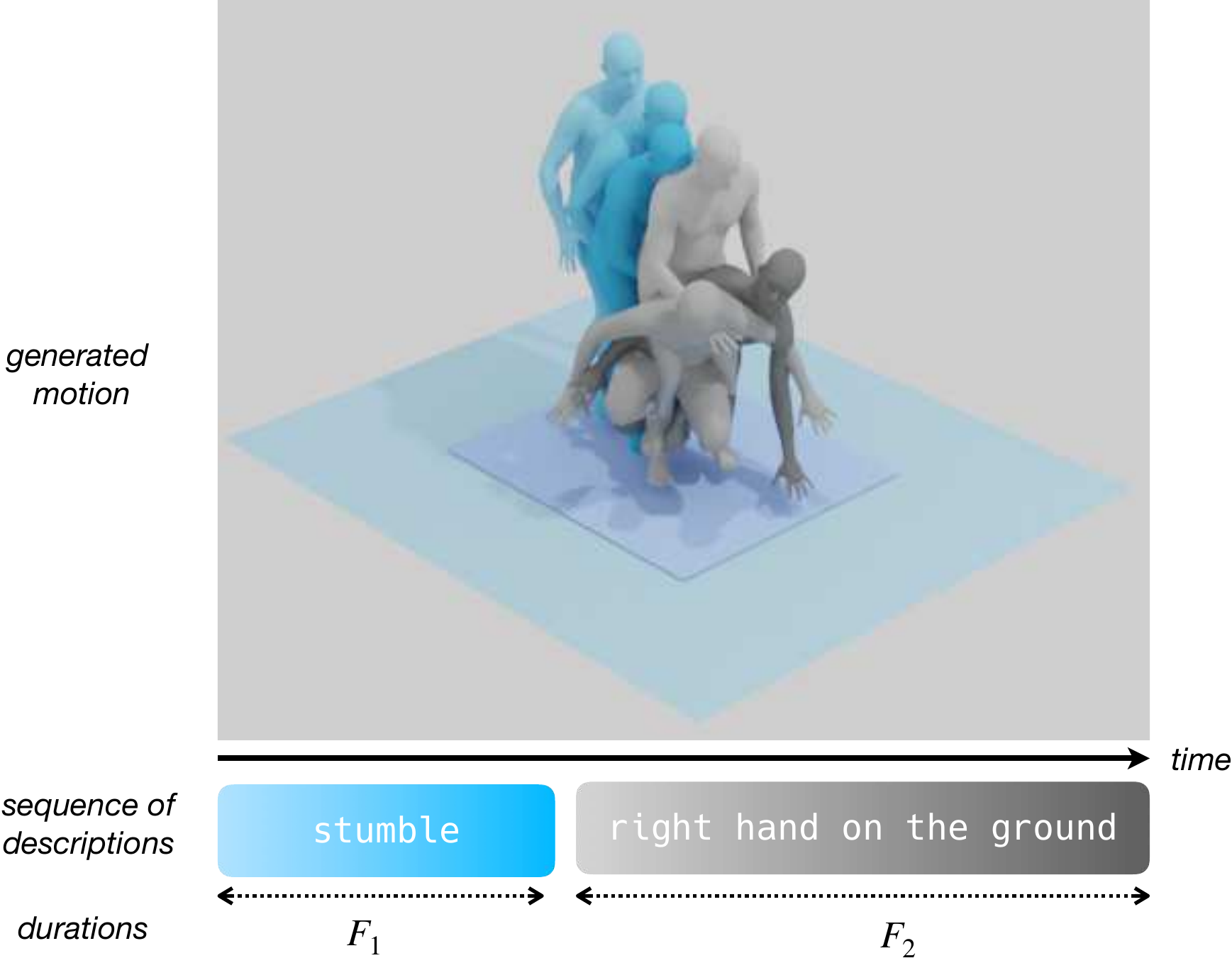}
	\vspace{0.1cm}
	\caption{\textbf{Goal:} Given a sequence of descriptions and durations as input, our goal is to generate a 3D human motion respecting the instruction and achieving temporal action compositionality. We design a recursive approach, \methodname, that can produce a variable number of actions given a stream of textual prompts. Note that the color saturation is aligned with the progress of each action.}
	\label{fig:teaser}
\end{figure}
Generative models are popular for synthesizing images
conditioned on textual descriptions~\cite{dalle2, Imagen}. Their success can be largely
attributed to massive training datasets. The same breakthrough has not
happened for 3D human motion generation due to lack of motion-text data.
Standard benchmarks for the text-conditioned motion synthesis task (e.g., the KIT Motion-Language dataset~\cite{Plappert2016_KIT_ML}) are limited in the vocabulary of actions and the number of motion sequences. In this work, we use the recently released
textual annotations of the BABEL dataset~\cite{BABEL:CVPR:2021}, 
providing English descriptions for the AMASS motion capture
collection~\cite{AMASS:ICCV:2019}.
This dataset is both larger and more diverse than previous datasets.
While previous work uses BABEL for its categorical
action annotations ($60$/$120$ classes) focusing mostly on classification settings \cite{tevet2022motionclip} or motion generation~\cite{song2022actformer},
we directly train with its free-form language descriptions, which have
not been used before. Thus, our generated motions
cover a significantly wider variety of actions compared to the
state of the art \cite{petrovich2022temos}.

Recently, TEMOS~\cite{petrovich2022temos} established
a new baseline in text-conditioned 3D motion synthesis
using Transformer-based VAEs~\cite{ACTOR:ICCV:2021}
and pretrained language models~\cite{distilbert_sanh}
to sample realistic motions corresponding to a text input.
This model is limited in several ways besides
being trained on the KIT data. 
TEMOS is not directly applicable for our task of generating a \textit{sequence}
of actions. 
While its non-autoregressive formulation
generates high-quality motions, the approach does not readily scale to long sequences of multiple motions due to the quadratic time complexity of Transformers.
Moreover, to embed complex sequences of actions in the latent space would require seeing a combinatorial number of action combinations during training.
With existing training data, generalization to new sequences would be challenging.
In our work, we combine the best of both worlds, by designing an iterative model that generates one motion per action at a time, by conditioning on the previous
motion. Within each iteration, we keep the non-autoregressive action generation approach, which probabilistically generates diverse and high-quality motions~(see Fig.~\ref{fig:teaser}).
We experimentally show that our iterative method compares favorably against baselines that jointly or independently generate pairs of actions in a single shot~(Fig.~\ref{fig:baselines}).

One of the key challenges in synthesizing long action sequences
given a  stream of textual prompts is how to ensure \textit{continuity}
within the transitions between actions. Independently generating one motion per action would not guarantee temporal smoothness.
In our framework, we find that encoding the next action conditioned on the last few frames of the previous action is a simple and effective solution.
To account for any remaining discontinuities still present with this solution,
we apply spherical linear interpolation (Slerp) over a short time window. 
Note that our approach treats transitions significantly better than a baseline that uses Slerp to interpolate between independently generated actions.

Our contributions are the following:
(i)~We introduce and establish a new benchmark for temporal action composition of 3D motions on the BABEL dataset;
(ii)~We design a new hybrid neural network model, \methodname (TEmporal Action Composition for 3D Humans), that addresses the limitations of previous state of the art by iteratively
generating infinitely many actions with smooth transitions;
(iii)~We obtain promising results for text-to-motion synthesis from a large-vocabulary of actions.
Our code and models will be available for research at \url{teach.is.tue.mpg.de}.
\section{Related Work}
There is a large and diverse body of work on motion synthesis including prediction~\cite{Zhang2020mojo}, infilling~\cite{Harvey2020RobustMI,Kaufmann:2020}, synthesis from action labels~\cite{chuan2020action2motion, ACTOR:ICCV:2021} or descriptions~\cite{petrovich2022temos}, and conditioned on constraints like poses~\cite{Kim2022ConditionalMI} or environment/objects~\cite{hassan_samp_2021,Starke2019NeuralSM}. First, we describe related work on motion prediction and synthesis briefly. Then, we focus on work that considers semantic control in various ways such as conditioning on action labels or language descriptions.

\noindent\textbf{Motion prediction.} %
Early work relies on statistical models to predict future frames of human motion, conditioned on past motion~\cite{Bowden2000LearningSM, Galata1999LearningSB,Galata2001LearningVM}, or synthesize cyclic motions and locomotion~\cite{Ormoneit2005RepresentingCH,Urtasun2007ModelingHL}. Due to the complex spatio-temporal nature of the input data, commonly modeled using skeletal representations, most recent work relies on autoregressive neural models~\cite{Fragkiadaki:2015,Gopalakrishnan2019ANT, Martinez:2017,Pavllo:2019}. Such models can process variable sequence lengths. 
On the other hand, other methods generate an entire sequence at once through convolution~\cite{Holden:2016,Yan2019ConvolutionalSG}, thus being more computationally efficient but less controllable. The majority of aforementioned approaches are deterministic, while human motion is inherently highly stochastic and diverse. To address this, recent work uses probabilistic models~\cite{Barsoum:2017,Yuan2020DLow, Zhang2020mojo, Zhao2020BayesianAH}.
Similarly, DLow focuses on strengthening the diversity of samples generated from a pretrained model~\cite{Yuan2020DLow}, while MOJO extends this to exploit the full 3D body~\cite{Zhang2020mojo}.

\noindent\textbf{Motion in-filling.} 
An alternative to motion prediction is to generate individual actions or poses and then fill in the transitions between them -- sometimes called in-betweening. Initial attempts at transition generation based on keyframes relied on inverse kinematics (IK) and time constraints to maintain physically-plausible motion~\cite{Rose1996EfficientGO, Witkin1988SpacetimeC}.
Recent approaches learn more expressive transitions from data \cite{Duan2021SingleShotMC,rtn,Harvey2020RobustMI,Ruiz2019HumanMP,2018-MIG-autoComplete,Zhou:2020}. 
Zhang~\etal~\cite{2018-MIG-autoComplete} use an RNN to learn jumping motions of a 2D lamp, while Ruiz~\etal~\cite{Ruiz2019HumanMP} and Kaufmann~\etal~\cite{Kaufmann:2020}, inspired by image inpainting, tackle motion infilling with CNNs, which work well for such tasks in 2D. 
Harvey~\etal propose Recurrent Transition Networks~\cite{rtn}, which are limited to fixed-length transitions, and later extend this to a stochastic model along with a benchmark for motion inbetweening~\cite{Harvey2020RobustMI}. Zhou~\etal~\cite{Zhou:2020} suggest that convolutions are better suited to the task of transition generation and, thus, propose a purely convolutional architecture with separate components for path predictor, motion generator and discriminator. Their method can interpolate over long time periods, but the maximum interval is limited by the receptive field of the generator. 
Most of this prior work on infilling and synthesis is focused on locomotion (walking, running, etc.).
While such tasks are important, human behavior includes a wider range of actions and their complex combinations.

\noindent\textbf{Conditioned motion generation.}
All the above work does not consider semantics or enable control of the motion.
Animating human skeletons or bodies using semantic control has a long history \cite{Badler1993SimulatingHC}. Recently, Holden~\etal propose PFNN~\cite{Holden2017PhasefunctionedNN}, which uses phase variables based on contact estimation to control the generation, while Zhang~\etal~\cite{Zhang:2018} use a mixture-of-experts scheme to dynamically compute the architecture weights. Starke~\etal extend the idea to include scene constraints \cite{Starke2019NeuralSM} and interactions of different body parts with objects \cite{Starke:2020b}. 
MotionVAE~\cite{Ling:2020} learns a human motion model using a VAE and then uses deep reinforcement learning to produce high-quality goal-directed motion. 
However, the above work requires labor-intensive data augmentation and is limited to a small number of actions, scenes and contexts, limiting its generalization.

More relevant here is work that generates human motion conditioned on language~\cite{Ahn2018Text2ActionGA,Ahuja2019Language2PoseNL,chuan2020action2motion,ACTOR:ICCV:2021, petrovich2022temos} or music~\cite{VallePerez:2021, Li:2021}. 
Text2Action~\cite{Ahn2018Text2ActionGA} relies on an encoder-decoder RNN to learn the mapping between language and pose. Language2Pose~\cite{Ahuja2019Language2PoseNL} instead learns a joint embedding space of sentences and poses, while Ghosh~\etal~\cite{Ghosh_2021_ICCV} synthesize 3D skeletons using sentences from the KIT dataset, encoding the upper and lower body separately. 
Action2Motion~\cite{chuan2020action2motion} and ACTOR~\cite{ACTOR:ICCV:2021} are two recent approaches that deal with the problem of motion generation conditioned on action labels. 
Action2Motion uses a GRU-VAE that operates on a per-frame basis. ACTOR~\cite{ACTOR:ICCV:2021}, on the other hand, employs a transformer-VAE that encodes and decodes motion sequences in one-shot. However, both approaches focus on a relatively small set of action labels, while we demonstrate success on a larger set of diverse actions using natural language descriptions. Concurrent to our work, Mao~\etal~\cite{mao2022weakly} propose a method for weakly supervised action-driven motion prediction. Given a set of frames of one action they generate the next action and the transition to it, using $20$ action categories from BABEL that have clear transitions.
In contrast, we perform motion generation directly from free-form text, going beyond categorical actions.

Most closely related to our work is TEMOS~\cite{petrovich2022temos}, which uses a Transformer-VAE and generates motions conditioned on free-form text inputs that are encoded using DistilBERT~\cite{distilbert_sanh}.
In contrast to that work, we use a larger and more challenging dataset~\cite{BABEL:CVPR:2021} that contains more diverse actions compared to KIT~\cite{Plappert2016_KIT_ML}, with precise, frame-level, annotations. 
We note that TEMOS cannot synthesize sequences of actions, and naively concatenating sequences of actions for training is difficult to scale both computationally and semantically to a large set of actions. 
Natural human motions are independent but also constrained when performed sequentially. Based on this observation, we introduce TEACH, which explicitly deals with sequences of actions, a unique property present in BABEL, generating actions from text prompts conditioned on the previous one.

\section{Motion Synthesis with \methodname}
\label{sec:method}

\begin{figure}
	\includegraphics[width=0.99\linewidth]{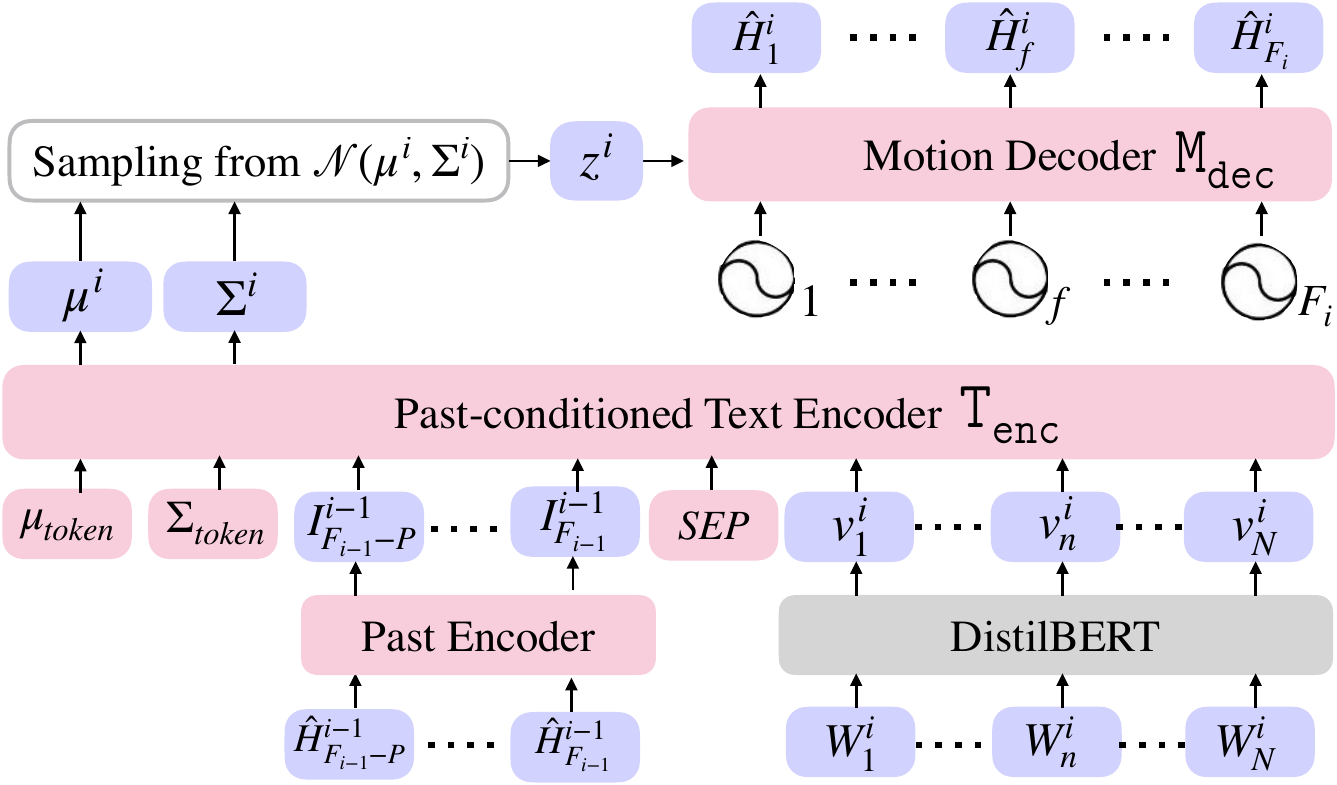}
	\caption{\textbf{Method overview:} Our \methodname model is a variational encoder-decoder neural network. The current text instruction and the past frames are encoded by the corresponding encoders and are fed to $T_{enc}$ along with the additional tokens. $T_{enc}$  produces the distribution parameters from which the latent vector is sampled and given to the decoder to generate a sequence of 3D human poses. In this figure, we omit the motion encoder for simplicity.}
		\label{fig:method}
\end{figure}

After defining the task (Sec.~\ref{subsec:task}), we present the TEACH architecture and explain the different baselines and architectural components of our method (Sec.~\ref{subsec:arch}).
Finally, we describe the training procedure and the losses (Sec.~\ref{subsec:training}).

\subsection{Task definition}
\label{subsec:task}

Starting from a sequence of instructions in natural language (e.g., in English), our goal is to generate smooth and realistic human motions that correspond to the instructions. 
Here, we demonstrate results for pairs or triplets of actions but our model can autoregressively generate an arbitrary sequence of actions given the respective action descriptions. During training, TEACH takes as input: a sequence of English prompts $S_1, \ldots, S_K$, where each phrase corresponds to an action description and consists of a sequence of words $S_i = W_1^i, \ldots, W^i_N$, \eg ``turn to the right'', ``walk forward'', ``sit down'', etc.~and a 3D human motion sequence. Each sequence consists of poses, $H_1, \ldots, H_F$, parametrized by the SMPL body model~\cite{SMPL:2015}. In this work, we follow the representation described in~\cite{petrovich2022temos} that converts SMPL parameters to a $6D$ rotation representation~\cite{zhou2019continuity} together with root translation. Moreover, we use the same normalization and canonicalization process as in~\cite{petrovich2022temos}.

\subsection{Architecture}
\label{subsec:arch}

Our architecture is inspired by TEMOS~\cite{petrovich2022temos}. We use the same language encoder~(DistillBERT) and motion encoder. We omit the details of the motion encoder as it is the same used in TEMOS. However, TEMOS is constrained to output a sampled motion given a language description, without being able to handle sequences of actions. 
Thus, we design a new text encoder architecture, which includes a Past Encoder~(PC) that provides our method with the context of the previous action when generating the second action in each pair. 
For the first motion in each pair, we disable the Past Encoder and only use the learnable tokens and the encoded text. A separation token is used to facilitate disambiguation of motion and text modalities in the model~\cite{devlin-etal-2019-bert}. 
As illustrated in Fig.~\ref{fig:method}, we encode the current text instructions $W_1^i, \ldots, W_N^i$ using a pre-trained, frozen text model (DistilBERT) into text features $\mathrm{v}_1^i, \ldots, \mathrm{v}_N^i$. Moreover, the last $P$ frames of the previous generated motion, $\widehat{H}^{i-1}_{F_{i-1}-P:F_{i-1}}$, are encoded into motion features $I_{F_{i-1}-P:F_{i-1}}$ (Past Encoder). Then, we combine the features from the previous action, $I^{i-1}_{F_{i-1}-j}, j \in \mathbb{N}$, and the current text features along with learnable tokens ($\mu_\text{token}$, $\Sigma_\text{token}$ and $\mathit{SEP}$), and pass them as inputs to the Past-conditioned Text-Encoder, which generates the distribution parameters $\mu^i$ and $\Sigma^i$. $\mu^i$ and $\Sigma^i$ are treated as parameters of a Gaussian distribution, from which we sample and decode the final motion. Next we explain each module separately.

\noindent\textbf{Past-conditioned Text Encoder.} We first encode the natural language descriptions with a frozen DistilBERT~\cite{distilbert_sanh} which takes as input the current text instruction $S_i = W^i_1, \ldots, W^i_N$ and outputs text features $\mathrm{v}^i_1, \ldots, \mathrm{v}^i_N$. We use a Transformer encoder architecture to encode the past motion corresponding to the last $P$ frames of the previous action. 
The past motion $\hat{H}^{i-1}_{F_{i-1}-P}, \ldots, \hat{H}^{i-1}_{F_{i-1}}$ is transformed into pose features ${I}^{i-1}_{F_{i-1}-P}, \ldots, {I}^{i-1}_{F_{i-1}}$.
Finally, we use a transformer encoder (the Past-conditioned Text Encoder module $T_{enc}$ to jointly encode  the past motion features and the current text features into $\mu^i$ and $\Sigma^i$, parameters of a Gaussian distribution. This network takes as extra inputs, the $\mu_{\text{token}}$ and $\Sigma_{\text{token}}$ as in ACTOR~\cite{ACTOR:ICCV:2021}, and a special token ($\mathit{SEP}$) to separate both modalities.
From the Gaussian distribution $\mathcal{N}(\mu^i, \Sigma^i)$, we sample a latent vector $z^i$.  For the first motion we disable PC since there is no previous motion.

\noindent\textbf{Motion decoder.} We use the same decoder architecture as in TEMOS~\cite{petrovich2022temos}, which generates a sequence of poses from a single embedding. This Transformer-based motion decoder takes the current latent vector $z^i$ and $F_i$ positional encodings (in the form of sinusoidal functions) as input, and generates the sequence of human motions.

\noindent\textbf{Baselines.} As there is no prior work that explicitly deals with sequences of actions, we design several baselines using  TEMOS, Slerp~\cite{slerp}, and geometric transformations. Note that our work is different from the action-driven motion prediction proposed concurrently in~\cite{mao2022weakly}, as we deal with full motion generation and free-form language descriptions and do not explicitly use only pairs formed by transitions. Specifically (Sec.~\ref{fig:baselines}), we employ two baselines: 
``Independent'', which is based on TEMOS and is trained on single action segments, and ``Joint'', which is also based on TEMOS, but takes as input the both motions (i.e., concatenation of the respective segments) and the corresponding language labels separated with a comma. 
For the case of Independent, the generated motions are fused into a pair of actions by: (1) aligning the last generated frame of the first action with the first frame of the second action by orientation and translation and (2) applying spherical interpolation to fill in the remaining transition between the two actions.

\subsection{Training}
\label{subsec:training}
\noindent\textbf{Data handling.}~BABEL consists of language descriptions and action categories for the majority of sequences in AMASS~\cite{AMASS:ICCV:2019}. Each sequence is separated in segments that can overlap without any constraint, except that all the frames of the sequence must be labeled. 
To train the Independent baseline, we use the training set segments from BABEL. 
Furthermore, we extract pairs of actions to train the remaining models. 
To achieve this, we process each sequence and, for each segment, $s_i = [t_s^i, t_e^i]$, we calculate all the segments $s_j$ : ${s_j \cap s_i \neq \varnothing , s_j \nsubseteq s_i, s_j \nsupseteq s_i }$,
except if a segment is a ``transition". 
We think of transitions happening {\em between} actions so a transition and an action cannot form a pair.
Simply, we use all the segments that are not a superset or subset of each other and have an overlap. Next, if a segment is connected to another segment via a transition (i.e., the same transition overlaps with both), that triple is considered a pair of actions. For the rest of the cases, the pairs are formed by the segment overlaps and not transitions.

A training iteration consists of two forward passes. The first action, corresponding to sentence description $S_1$, and the given length $F_1$, will produce $\mu^1$, $\Sigma^1$ and the generated motion $\hat{H}^1_1, \ldots \hat{H}^1_{F_1}$. Then, given the second instruction $S_2$, the given length $F_2$, and the $P$ last frames of the previous generated motion, we produce $\mu^2$, $\Sigma^2$ and the generated motion of the second segment $\hat{H}^2_1, \ldots \hat{H}^2_{F_2}$. We do one backward pass, which  optimizes the reconstruction loss and the KL loss on the two segments jointly.

\noindent\textbf{Reconstruction loss.}~From the two forward passes, we generate the motions $\widehat{H}^{1}_{1:F_1}$ and $\widehat{H}^{2}_{1:F_2}$. We enforce them to be close to the corresponding ground truth motions $H^1_{1:F_1}$ and $H^2_{1:F_2}$ via the following loss terms:
\begin{equation} \label{eq:lr}
\mathcal{L}_{R} = \mathscr{L}(H^1_{1:F_1}, \widehat{H}^{1}_{1:F_1}) + \mathscr{L}(H^2_{1:F_2}, \widehat{H}^{2}_{1:F_2}), 
\end{equation} 
where $\mathscr{L}$ is the smooth L1 loss.

\noindent\textbf{KL loss.}~By using the notation $\phi^i = \mathcal{N}(\mu^i, \Sigma^i)$, and $\psi = \mathcal{N}(0, I)$, this loss regularizes the two Gaussian distributions $\phi^1$ and $\phi^2$ to be close to $\psi$ as in the VAE formulation. We minimize the two Kullback-Leibler (KL) divergences
\begin{equation} \label{eq:kl}
\mathcal{L}_{KL} = KL(\phi^1, \psi) + KL(\phi^2, \psi).
\end{equation}
We also use the same additional KL losses as TEMOS, to enforce the latent vectors to follow the same distributions and the same $L_1$ loss to keep them as close as possible. We omit them from the description for simplicity and to highlight our technical contributions.
The total loss is a weighted sum of the two terms: 
$\mathcal{L} = \mathcal{L}_{R} + \lambda_{KL}\mathcal{L}_{KL}$. In practice, we  use $\lambda_{KL}=10^{-5}$ as in TEMOS~\cite{petrovich2022temos} and ACTOR~\cite{ACTOR:ICCV:2021}.

\noindent\textbf{Implementation details.}~For both the Past-Conditioned Text Encoder and the Past Encoder, we use a Transformer encoder model with $6$ layers and $6$ heads, a dropout of $0.1$ and a feed-forward size of $1024$. The latent vector dimension is 256. The whole model is trained with the AdamW optimizer~\cite{adamW} with a fixed learning rate of $10^{-4}$ with a batch size of $32$ or $16$. Both during training and test time, we use ground truth durations ($F_i$).
We also use Slerp~\cite{slerp} and alignment between the first and the second action for TEACH as well as for the Independent baseline. We apply Slerp to interpolate for $8$ frames at the beginning of the second motion which includes the transition.
\section{Experiments}
\label{sec:experiments}
We first describe the dataset (Section~\ref{subsec:data})
and evaluation metrics (Section~\ref{subsec:metrics}) used in our experiments.
We then report our main results by comparing our method 
with multiple baselines (Section~\ref{subsec:baselines}).
Next, we present an ablation study to 
investigate the contribution of motion interpolation (Section~\ref{subsec:slerp})
and different numbers of past frames used by PC (Section~\ref{subsec:past}).
Finally, we provide qualitative results (Section~\ref{subsec:qualitative})
and a discussion on limitations (Section~\ref{subsec:limitations}).

\begin{figure}
	\centering
	\includegraphics[width=0.49\textwidth]{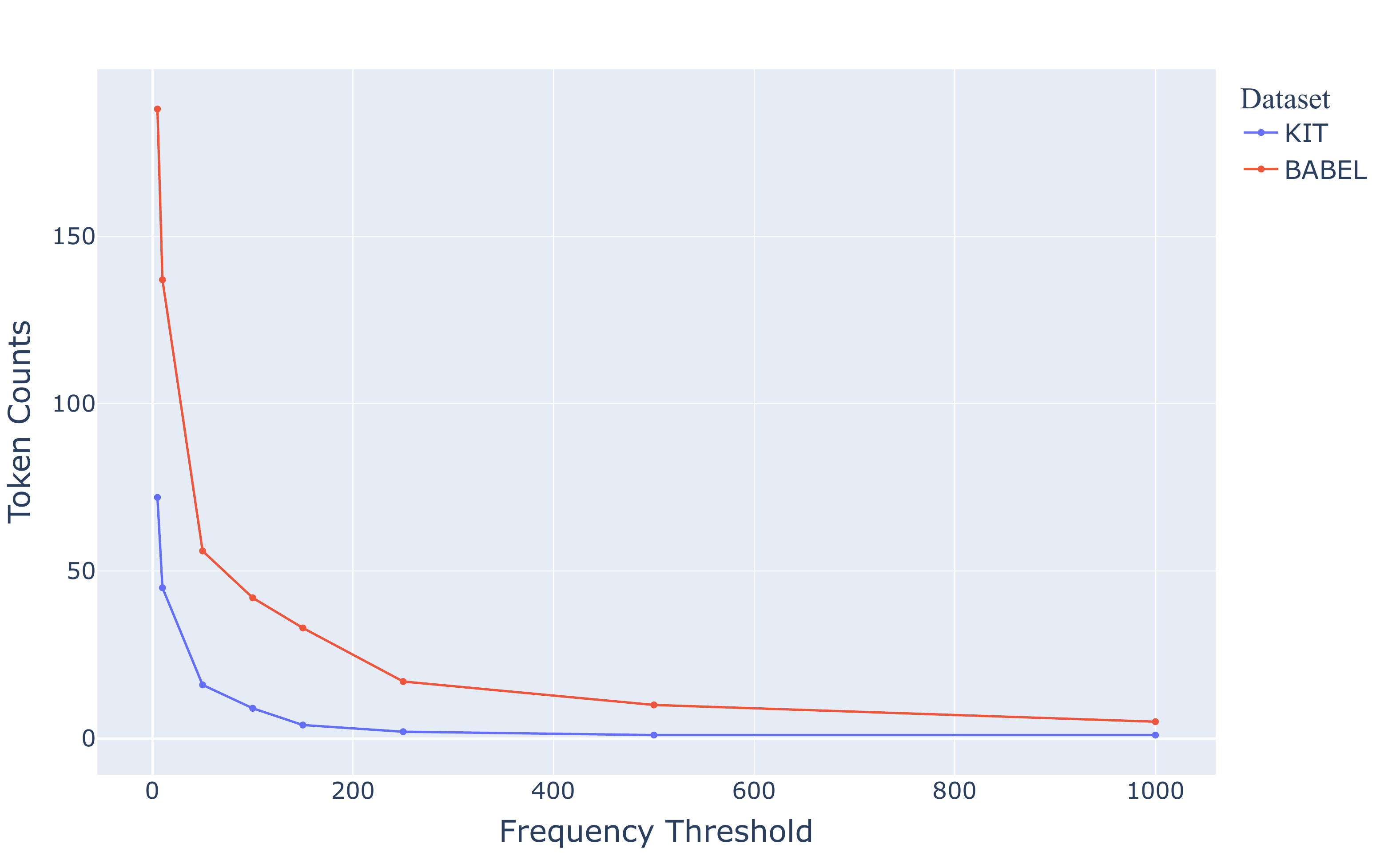}
	\includegraphics[width=0.49\textwidth]{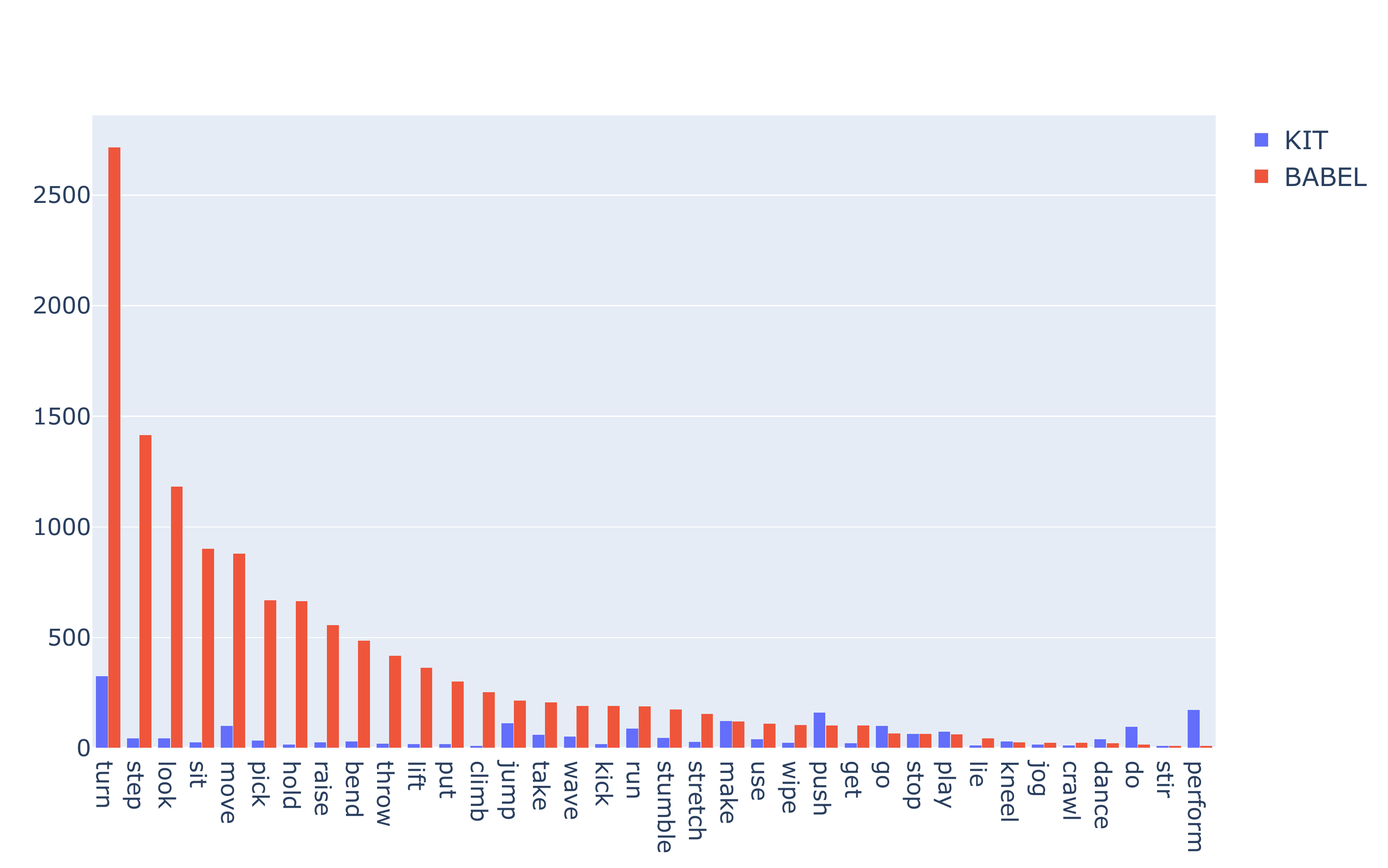}
	\caption{\textbf{BABEL vs KIT:} We provide a comparative analysis of the
	amount of data and the vocabulary of verbs. On the top, the number of tokens~(i.e. different words) in each dataset is plotted against various frequency thresholds, i.e. the number of words that appear at least freq.~threshold times. We see that BABEL consistently has at least twice as many tokens as KIT. On the bottom, the verb histogram shows that BABEL has more samples across a wide range of actions.
	Note that there are differences in how the datasets label actions with generic words like ``do" and ``perform" being common in KIT and rare in BABEL, which is more specific.
	}
	\label{fig:babel_stats}
\end{figure}

\subsection{BABEL dataset}
\label{subsec:data}
In our work, we train and evaluate on BABEL~\cite{BABEL:CVPR:2021},
which provides textual descriptions for the motions in the AMASS
collection~\cite{AMASS:ICCV:2019}. In particular, we use the
processed text version~(lemmatized etc.~as opposed
to the raw version which is also provided). %
We do not use the categorical action labels. In total there are $10881$ motion sequences, with $65926$ textual
labels and the corresponding segments. 
The unique property of BABEL is that it has annotated segments that overlap in each sequence, which allows us to investigate generation of a \textit{sequence of actions}.
In contrast, a textual label in KIT~\cite{Plappert2016_KIT_ML} covers the entire
sequence. Moreover, KIT is smaller both in terms of the number of motion sequences and the number of actions. 
Fig.~\ref{fig:babel_stats} shows  the distribution of verbs according to the most frequent verbs in BABEL. 
Refer to Sup.~Mat.~for additional analysis of KIT's most frequent verbs compared with BABEL
and also analysis about other part-of-speech categories.
There are approximately 5.7k and 23.4k pairs in the validation and training sets respectively. We consider pairs of actions for simplicity but \methodname is applicable to sequence of actions of arbitrary length. Note that, we do not use ``t-pose'' or ``a-pose'' actions during training. We use transitions only to identify possible pairs of actions. During training, in the case of segment overlap, we uniformly distribute the overlapping frames across the two segments that constitute the pair. Also, note that the majority of the pair data ($\sim 70\%$) is created by overlapping segments and not by transitions. In the case of a transition, we concatenate the transition with the second segment.
\begin{table*}[htbp]
    \centering
    \setlength{\tabcolsep}{4pt}
    \resizebox{0.99\linewidth}{!}{
    \begin{tabular}{l|rrrr|rrrr}
        \toprule
         \multirow{2}{*}{\textbf{Methods}} & \multicolumn{4}{c|}{Average Positional Error $\downarrow$} & \multicolumn{4}{c}{Average Variance Error $\downarrow$} \\
         & \small{root joint} & \small{global traj.} & \small{mean local} & \small{mean global} &
         \small{root joint} & \small{global traj.} & \small{mean local} & \small{mean global}  \\
    \midrule
    \textbf{(a) Independent} &  0.729 &  0.707 &   0.169 &     0.770 &  0.255 &  0.253 &   0.016 &     0.267 \\ %
    \textbf{(b) Joint} & 0.790 & 0.773 & 0.163 & 0.832 & 0.306 & 0.305 & 0.014 & 0.317 \\
    \textbf{(c) Past-conditioned (\methodname)} & 0.674 &  0.654 & 0.159 & 0.717 & 0.222 & 0.220 & 0.014 & 0.234 \\ %
    \bottomrule
    \end{tabular}
    }
    \vspace{0.2cm}
    \caption{\textbf{Comparison against baselines on pairs of actions:}
      We benchmark the 3 different approaches on pairs of BABEL~\cite{BABEL:CVPR:2021}. As we can see TEACH outperforms Joint and Independent baselines in all the metrics. 
    }
    \label{tab:baselines}
\end{table*}

\subsection{Evaluation metrics}
\label{subsec:metrics}

We follow the evaluation metrics employed by \cite{Ghosh_2021_ICCV,petrovich2022temos},
namely Average Positional Error (APE) and Average Variational Error (AVE),
measured on the root joint and the rest of the body joints separately. Mean \textit{local} and \textit{global} refer to the joint position in the local (with respect to the root) or global coordinate systems, respectively.
As in \cite{petrovich2022temos}, we sample one random motion generation from our variational model and compare it against the ground truth motion corresponding to the test description.
While we quantitatively evaluate on pairs of actions, %
we qualitatively show the ability of our model to generate two or more actions in Fig.~\ref{fig:qualitative} and the supplementary video.

\begin{figure}[t]
	\includegraphics[width=0.99\linewidth,trim={0 0 0 1},clip]{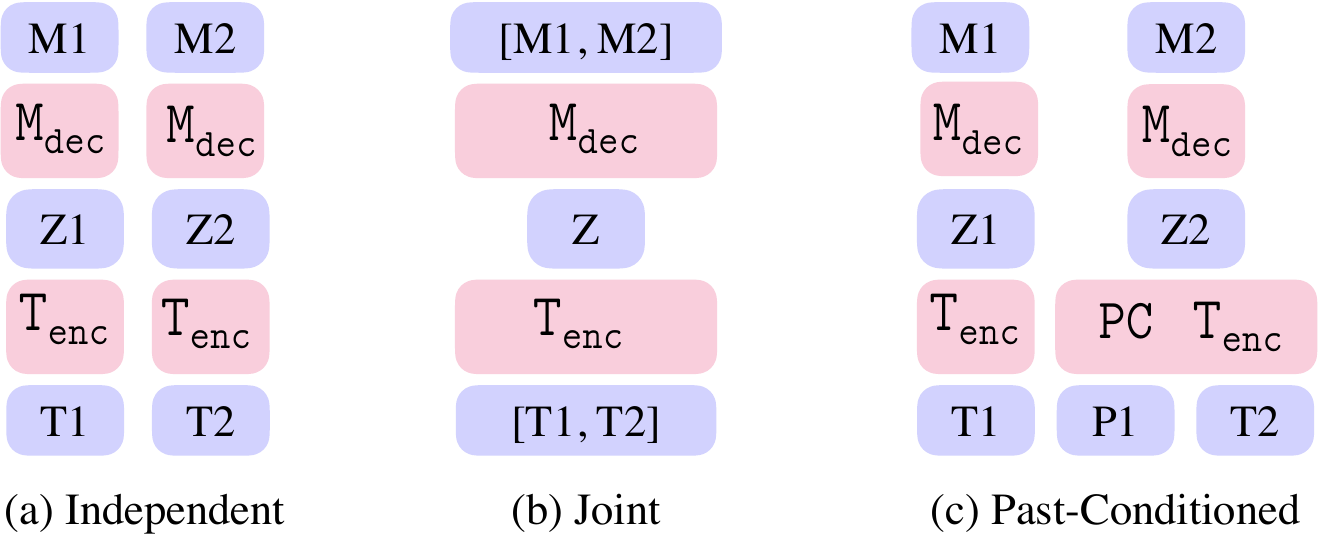}
	\caption{\textbf{Variants:} We illustrate the baselines for independent single-action training (a) and joint two-action training (b). Our method on the other hand is recursive, and is conditioned on the past motion (c). T1 and T2 denote the sequence of two textual descriptions. M stands for motion, and Z stands for the latent vector.}
	\label{fig:baselines}
\end{figure}

\begin{figure*}[htbp]
    \centering
    \includegraphics[width=\textwidth]{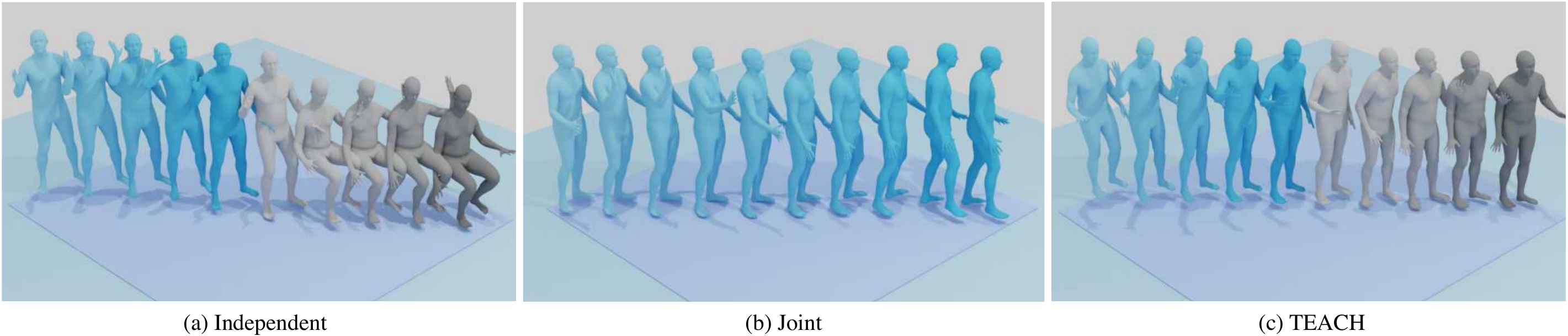}
    \caption{\textbf{Qualitative comparison:} We show an illustrative example for (a) Independent, (b) Joint and (c) TEACH %
    for the sequence of actions \texttt{[wave the right hand}, \texttt{raise the left hand]}.
    While the individual waving and raising hand actions are correctly generated, the single-action independent baseline (a) transitions from standing to sitting incoherently as the next action is not conditioned on the past.
    Joint baseline (b) on the other hand, waves with the right hand but does not raise the left one, probably because such an action combination was not present in the training set. On the other hand, TEACH learns about both single action variation and and autoregressive transitions between actions, and thus completes both actions naturally. Note that, while these motions are performed in place, we
    artificially translate each pose to show the motion frame-by-frame
    such that the transition and action details are easier to see.
    }
    \vspace{0.3cm}
    \label{fig:comparison}
\end{figure*}

\subsection{Comparison with baselines}
\label{subsec:baselines}

Here, we first describe the baselines we created
by adapting TEMOS~\cite{petrovich2022temos}
to the action sequence synthesis task without any architectural changes.
Fig.~\ref{fig:baselines} summarizes the two variants (a) independent
and (b) joint training.

Independent training, in Fig.~\ref{fig:baselines} (a), refers to inputting a single
text and outputting a single motion, as is the case for TEMOS. To adapt this model to a sequence of actions, we generate the two actions and perform an interpolation operation (i.e., Slerp) to obtain smooth transitions between the independently generated motions. However, a naive interpolation results in poor transitions since the second motion may start at a different
global location, with a different global rotation. To account for this mismatch, we translate the root of  the second motion to have the same x,y coordinates as the first motion. Then, we rotate it to match the first motion's global orientation. The advantage of this model is its ability to scale up to any number of action compositions. However, despite the interpolation, we observe unnatural motions due to large changes between body poses during
transitions. For example in Fig.~\ref{fig:comparison}, the model generates two motions that are not compatible in terms of pose, creating an unrealistic transition. This is expected as the independent baseline has no notion of the previous motions.

The joint training, in Fig.~\ref{fig:baselines} (b), is another alternative
to extend TEMOS to multiple descriptions without further modifications.
We simply combine the sequence of descriptions into a single text with a comma punctuation in between, and train the model with pairs of motions corresponding to consecutive actions. The advantage of this model is the ability to produce smooth motions, including the transitions. However, the major disadvantage concerns scalability. Due to quadratic complexity with respect to the motion duration, the joint training does not scale well to a large number of actions. Moreover, it would require many action combinations, i.e., a concatenation of more than pairs of actions, at training to produce a variable number of actions. Such data are not easy to capture, making it challenging to train such a model.
Finally, it may be difficult to generalize to unseen action combinations. In our experiments, we train this model with 2-action pairs (which is a relatively easy setting compared to more actions).

In contrast to independent and joint training, our model (Fig.~\ref{fig:baselines} (c)) is
recursive and the future action is conditioned on the previous action. In Tab.~\ref{tab:baselines}, we summarize the performance of these three variants on the BABEL validation set. Our past-conditioned \methodname, which uses the last $5$ frames from the previous action, outperforms the baselines. Due to the difficulty of quantitative evaluation of generative models, we also rely on qualitative comparisons, provided in the supplementary video. An illustration of our arguments can be seen in Fig.~\ref{fig:comparison}.

\begin{table*}[htbp]
    \centering
    \setlength{\tabcolsep}{4pt}
    \resizebox{0.99\linewidth}{!}{
    \begin{tabular}{l|cc|rrrr|rrrr}
        \toprule
         \multirow{2}{*}{\textbf{Methods}} & \multicolumn{2}{c|}{Transition Dist.} &  \multicolumn{4}{c|}{Average Positional Error $\downarrow$} & \multicolumn{4}{c}{Average Variance Error $\downarrow$} \\
         & w/ align. & w/out align. & \small{root joint} & \small{global traj.} & \small{mean local} & \small{mean global} &
         \small{root joint} & \small{global traj.} & \small{mean local} & \small{mean global}  \\
    \midrule
    \textbf{Independent (no Slerp)} & 0.151 & 0.177 & 0.762 &  0.740 & 0.170 & 0.805 &  0.255 &  0.253 & 0.016 & 0.267 \\
    \textbf{\methodname (no Slerp)} & 0.107 & 0.122 & 0.677 & 0.658 & 0.159 & 0.722 &	0.227 & 0.225 &	0.015 & 0.239 \\
    \textbf{Independent}            & n/a   & n/a   & 0.729 &  0.707 & 0.169 & 0.770 &  0.255 &  0.253 & 0.016 & 0.267 \\
    \textbf{\methodname}            & n/a   & n/a   & 0.674 &  0.654 & 0.159 & 0.717 & 0.222 & 0.220 & 0.014 & 0.234\\     
    \bottomrule
    \end{tabular}
    }
    \vspace{0.2cm}
    \caption{\textbf{Effect of Slerp:} We measure transition distance for generated samples given all the test set pairs. We define transition distance as the Euclidean distance between the last frame of the first action and the first frame of the second action, calculated on joint positions, when the last pose of the first action is aligned with the first pose of the next action and when it is not. TEACH better captures the transition between the two actions compared to the previous-action-agnostic TEMOS. Moreover, the Independent baseline cannot be benchmarked without orienting and aligning the poses as it is trained on single actions that are canonicalized to face in the forward direction.
    }
    \label{tab:slerp}
\end{table*}

\subsection{Effect of interpolating the action transitions}
\label{subsec:slerp}

As explained in Sections~\ref{sec:method}~and~\ref{subsec:baselines}, we use Slerp interpolation between actions both for the independent training baseline, and our method. We justify the use of such interpolation with the experiment in Tab.~\ref{tab:slerp}. %
Removing Slerp causes discontinuities which are easier to see
visually from our supplementary video. However, the discontinuity is higher for the independent generations than in \methodname.
To measure the degree of discontinuities, we report the average transition distance, i.e., the Euclidean distance between the two body poses corresponding to the last frame of the previous action, and the first frame of the next action. We see a clear decrease in discontinuity with \methodname (0.107 vs 0.151 m), even when the bodies are aligned. Moreover, we measure the same metric in the absence of alignment for the global orientation of the second motion (see Sec.~\ref{subsec:baselines}). We see that this alignment step is crucial for the independent baseline, as the transition distance compared to \methodname is even worse if we do not apply any alignment at all~(0.177 to 0.122), demonstrating that \methodname models generate smoother transitions than the baseline.

\begin{table}%
    \setlength{\tabcolsep}{2pt}
    
    \resizebox{0.99\columnwidth}{!}{%
    \begin{tabular}{l|rrrr|rrrr}
        \toprule
         \multirow{2}{*}{\textbf{$P$}} & \multicolumn{4}{c|}{Average Positional Error $\downarrow$} & \multicolumn{4}{c}{Average Variance Error $\downarrow$} \\
         & \small{root joint} & \small{global traj.} & \small{mean local} & \small{mean global} &
         \small{root joint} & \small{global traj.} & \small{mean local} & \small{mean global}  \\
    \midrule
     1  &  0.725 &  0.704 & 0.160 & 0.766 &  0.222 &  0.220 &   0.015 & 0.234 \\
     5 &  0.674 &  0.654 & 0.159 & 0.717 &  0.222 &  0.220 &   0.015 & 0.234 \\
     10 & 0.718 &  0.698 & 0.157 & 0.759 &  0.238 &  0.237 &   0.015 & 0.250 \\
     15 &  0.719 &  0.699 & 0.163 & 0.761 &  0.238 &  0.236 &   0.014 & 0.250 \\ 
    \bottomrule
    \end{tabular}

    \vspace{0.2cm}
}
\caption{\small \textbf{Ablation on the number of past frames:} Here, we change the number of past frame, while keeping the other training settings identical and report the different metrics. We observe the best performance when using $5$ past frames.
}

    \label{tab:past}
\end{table}

\subsection{Past conditioning duration}
\label{subsec:past}

In Tab.~\ref{tab:past}, we investigate
the influence of the hyperparameter $P$, the number of frames from the past motion to input to the past-conditioned text encoder.
While the performance is similar across $1$, $5$, $10$, or $15$ frames,
we observe a slight improvement when using $5$ frames as opposed to $1$ frame, potentially because a single frame does not capture enough past information. However,
further increasing the number of past frames
does not improve the results.

\subsection{Qualitative analysis}
\label{subsec:qualitative}

\begin{figure*}[t]
    \centering
    \includegraphics[width=\textwidth]{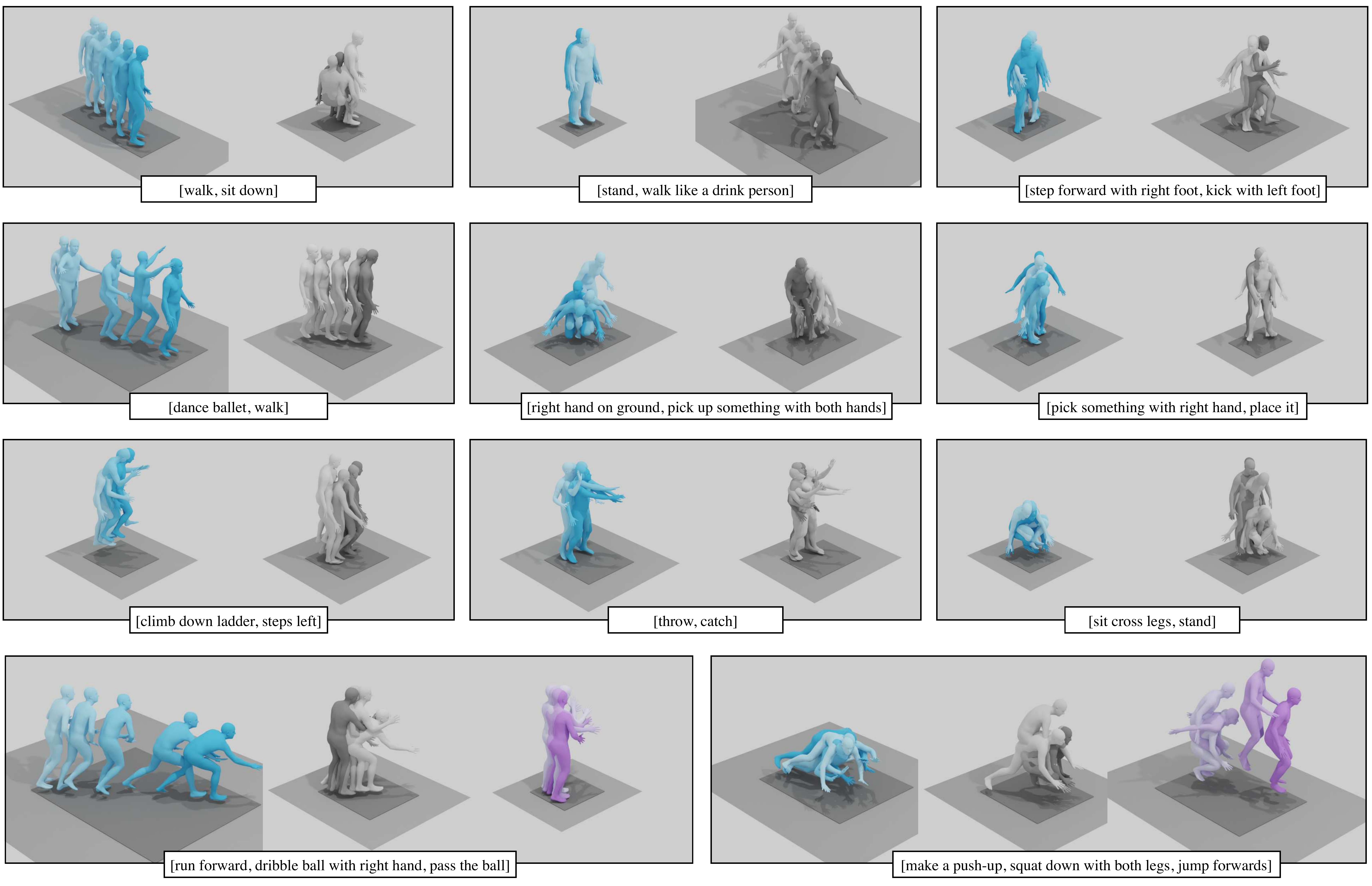}
	\caption{\textbf{Qualitative results:} In the first 3 rows, we visualize TEACH results for pairs of actions. We see how TEACH goes beyond walking and that in all the cases there are two actions being performed. Even fine-grained sequences of action like `step forward with the right foot' and `kick with the left foot' are generated accurately. In the final row, we show triplets of actions. We use a separate image for each action in the sequence to make the performed action clearly visible. We denote the ending of the first action with the most saturated version of cyan, while the starting of the second is the less saturated version of gray.
    }
    \vspace{0.3cm}
    \label{fig:qualitative}
\end{figure*}

We present qualitative motion generation results in Fig.~\ref{fig:qualitative}. In contrast to previous
work that trains models on the KIT dataset~\cite{Plappert2016_KIT_ML},
our model is able to go beyond locomotive motions,
and covers a wider variety of actions, such as \texttt{right hand on the ground}. Finally, we show examples of more than 2 actions in the last row of Fig.~\ref{fig:qualitative}.
We refer to the supplementary materials for viewing the motions
as a video, providing analyses of the effect of interpolation
and
motions beyond pairs of actions. 

\subsection{Limitations}
\label{subsec:limitations}
Our work does not come without limitations. TEACH is susceptible to acceleration peaks when transitioning from the first action to the second one. There is still the need to apply Slerp to smooth out these discontinuities but, as we see in Tab.~\ref{tab:slerp}, the starting/ending poses of the two actions are not far away. This behavior may be also attributed to the variational nature of the model, which makes it difficult to precisely match the previous motion without any explicit pose-level autoregressive constraints. Moreover, BABEL has a lot of overlapping segments of actions which makes it difficult sometimes to have a visually ``clear'' sequence of actions, as some actions might mix with others. 

\section{Conclusions}
\label{sec:conclusions}

We presented a new task on motion generation from
a sequence of textual prompts, which we refer to as action
compositions in time. We established a new benchmark
on the BABEL dataset for this task, and explored
a variety of strong baselines, including independently or jointly training pairs of actions. Our recursive approach, \methodname, improves over the baselines
quantitatively, while addressing the past limitations
by allowing variable numbers of actions and producing fewer
discontinuities at transitions.
While we obtain promising results within this new direction, there is still room for improvement. Motion realism can be improved and contact with the world could be explicitly modeled. Here, we assume that the character does not know what it will do in the future; that is, it only looks backwards in time. 
In contrast, humans have goals and know what they will do next. This knowledge about the future can affect the present.
Future work can explore such ``looking ahead'' to better generate realistic sequences of actions.
We hope that TEACH will encourage further research on combining language and 3D motion, much like the field has done with language and 2D images \cite{dalle2, Imagen}.

\bigskip
\noindent\textbf{Acknowledgements.}~We thank Benjamin Pellkopfer for his IT support and Vassilis Choutas for his suggestions. This work was granted access to the HPC resources of IDRIS under the allocation 2022-AD011012129R1 made by GENCI. GV acknowledges the ANR project CorVis ANR-21-CE23-0003-01 and the Google research gift. \textbf{Disclosure:}~\url{https://files.is.tue.mpg.de/black/CoI_3DV_2022.txt}

{\small
\bibliographystyle{ieee_fullname}
\bibliography{references}
}

\newpage
{\noindent \large \bf {APPENDIX}}\\
\renewcommand{\thefigure}{A.\arabic{figure}} %
\setcounter{figure}{0} 
\renewcommand{\thetable}{A.\arabic{table}}
\setcounter{table}{0} 

\appendix

In this document, we provide a description
of our supplementary video
(Section~\ref{sec:app:video}),
details on runtime (Section~\ref{sec:app:runtime}),
dataset statistics (Section~\ref{sec:app:langstats}),
and implementation details (Section~\ref{sec:app:implementation}).

\section{Supplemental video}
\label{sec:app:video}

We provide for the reader, an explanatory video that: 
(i) gives a quick overview of our method and baselines;
(ii) explains visually how the alignment of the different actions is performed;
(iii) more importantly, includes videos of generated motions for the different methods which are critical to assess the performance.
We show generated motions to provide
(a) comparisons between the ground truth, Independent, Joint and TEACH; %
(b) comparisons of TEACH and Independent without the use of Slerp;
(c) sequences of multiple actions for TEACH, going beyond pairs of actions;
(d) failures cases.

\section{Runtime}
\label{sec:app:runtime}

As described in
\if\sepappendix1{Section~3}
\else{Section~\ref{sec:method}}
\fi
of the main paper, we train $3$ models: Independent, Joint and TEACH. Our computational resources are various types of GPUs, mainly ``Tesla V100-PCIE-16GB'', ``Tesla V100-PCIE-32GB'', ``Tesla V100-SXM2-32GB'', ``Quadro RTX 6000-24GB''. We trained each model with a single GPU. Independent training time is 2 days approximately, Joint is 5 days and TEACH is 2 days to reach 600 epochs.
The Joint baseline due to the quadratic increase in time complexity for transformers is significantly slower to train and difficult to scale to more than two actions.

\section{Dataset statistics}
\label{sec:app:langstats}
Similar to
\if\sepappendix1{Section~4.1}
\else{Section~\ref{subsec:data}}
\fi
of the main paper,
we provide additional statistics on the BABEL dataset~\cite{BABEL:CVPR:2021}.

\begin{figure*}
	\centering

	\includegraphics[width=0.99\textwidth, trim={0cm 19cm 0cm 0cm}, clip]{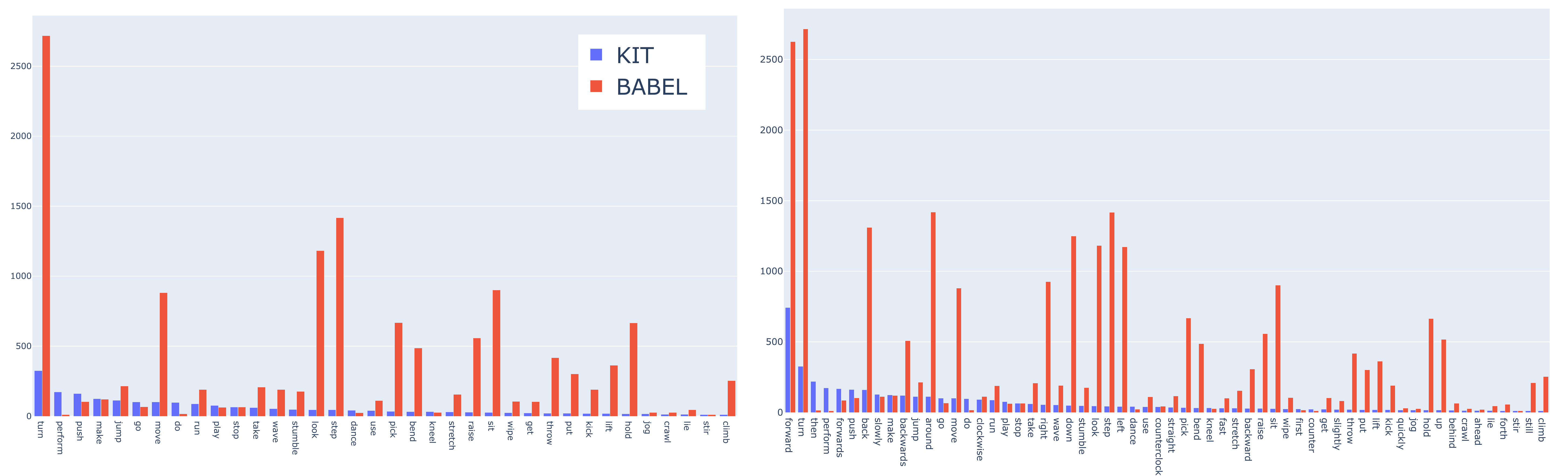}
	\includegraphics[width=0.99\textwidth, trim={6cm 0cm 12cm 0cm}, clip]{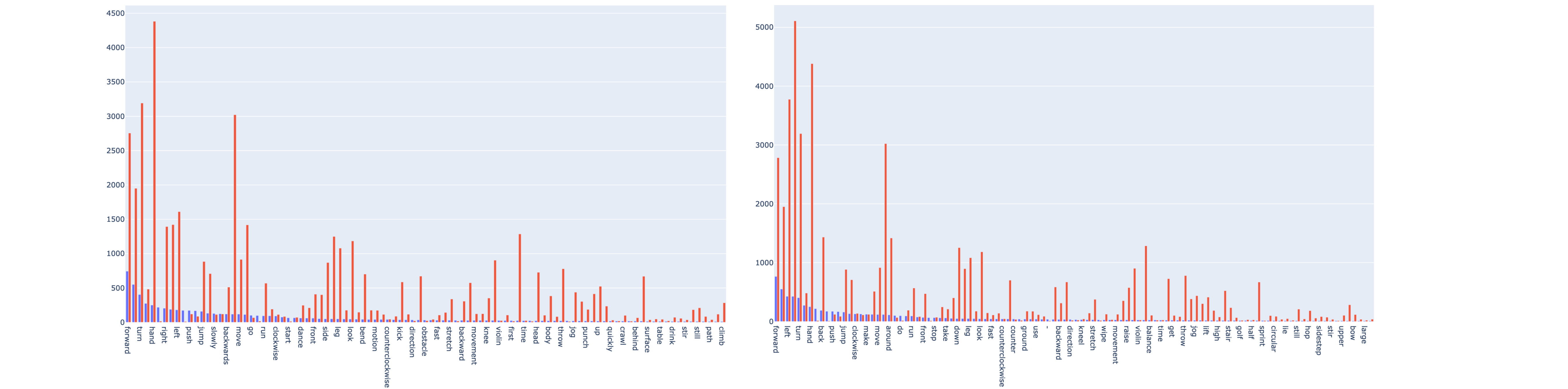}
	
	\caption{\textbf{BABEL vs KIT:} Here, we show additional plots regarding the language on BABEL and KIT. We show the token frequency of the two different datasets for different POS tag groups. Verbs~(top left), verbs and adverbs~(top right), verbs, adverbs and nouns (bottom left) and verbs, adverbs, noun, adjectives~(bottom right).
	}
	\label{fig:babel_stats_supmat}
\end{figure*}

\subsection{Language statistics}

In Fig.~\ref{fig:babel_stats_supmat}, we demonstrate the frequency of different tokens for the two datasets, but this time sorted according to the most frequent words in the vocabulary of the KIT dataset~\cite{Plappert2016_KIT_ML}. As we have shown in the main paper, BABEL is at least twice as rich in terms of language. Here, we show that, even for the most frequent tokens of KIT and for all the different POS (part of speech) tag combinations, the same tokens appear much more often in BABEL. Note that we show $4$ different plots for the $4$ cases of POS tags which are highly probable to involve an action.

\subsection{Duration statistics}
In Tab.~\ref{tab:stats-supmat}, we show the statistics of different data types in BABEL. To be clear, both Segments and Sequences datatypes are included in BABEL. Using the consecutive Segments, we build the pairs, as the Sequences contain only a single action. However, Sequences are included in the independent model training. We observe from Tab.~\ref{tab:stats-supmat}, that the mean length of the Segments is 3 times smaller than those of the Sequences. Moreover, the median is consistently smaller than the mean which implies that the distribution of the durations are long-tailed.
\begin{table}
    \begin{tabular}{@{}lrrrr@{}}
    \toprule
    Datatype   & Mean(s) & Std(s) & Median(s) & Samples(\#) \\ \midrule
    Segments   & 1.97    & 3.73   & 1.1       & 61639       \\
    Segments*  & 2.49    & 4.46   & 1.38      & 40395       \\
    Sequences  & 6.61    & 6.85   & 4.23      & 4287        \\
    Seg* + Seq & 2.88    & 4.89   & 1.52      & 44682       \\
    Pairs      & 4.98    & 5.98   & 3.4       & 49820       \\ \bottomrule
    \end{tabular}
    \caption{\textbf{Datatype statistics:}
    We show the statistics from different BABEL label types. Sequences are the AMASS~\cite{AMASS:ICCV:2019} motions with a single action label, Segments are the smaller motions that are extracted from longer AMASS sequences. Pairs are the two-action motions that we extract from consecutive segments. We denote the exclusion of ``transition'', ``a-pose'', and ``t-pose'' labels with *.}
    \label{tab:stats-supmat}
\end{table}

\section{Implementation Details}
\label{sec:app:implementation}

\subsection{Data processing}
We use BABEL's original splits for the training, validation set. We report our final results in the validation set, for easier reproduction, since BABEL test set is not publicly available. We use AMASS motions subsampled at $30$ fps. When training with pairs, we remove motion pairs that have a duration smaller than $0.3$ seconds or bigger than $25$ seconds. Similarly, we remove motion segments or sequences that are smaller than $0.3$ seconds when training the Independent model and whenever a motion is longer than $5$ seconds, we take a random $5$-second subset to feed as input. We train the independent baseline using both Segments and Sequences. %

For the canonicalization of input motions, we follow the same setup as in~\cite{petrovich2022temos}
by rotating the bodies to face the forward direction.
We canonicalize each action separately for the independent baseline.
For TEACH and the joint baseline, we canonicalize the entire sequence according to the first frame.
We also standardize the data (similar to \cite{petrovich2022temos}) for each of the different cases~(pairs, single-action). 

\subsection{Alignment \& interpolation}

For the case of Slerp and alignment, we translate and rotate the second motion such that the last frame of the first action and the first frame of the second action match.
Then, we interpolate between those aligned poses via Slerp.
We insert $8$ frames between the two actions for interpolation.
For the independent baseline, the alignment step is crucial since the single actions are independently generated without ensuring continuity within transition poses.

\end{document}